\documentclass[journal]{IEEEtran}

\usepackage{cite}

\usepackage{graphicx}
\usepackage{epstopdf}
\usepackage{stfloats}

\usepackage{multicol}
\usepackage{multirow}

\usepackage{amssymb}
\usepackage{amsmath}
\usepackage{array}
\usepackage{algorithm}
\usepackage{algpseudocode}

\graphicspath{{figs/}}

\usepackage{url}

\begin{document}
\hyphenation{op-tical net-works semi-conduc-tor}

%
\title{ChineseFoodNet: A Large-scale Image Dataset for Chinese Food Recognition}
%
%
%

\author{Xin~Chen$^{\dag}$,
        Yu Zhu$^{\dag}$,
        Hua~Zhou,
        Liang~Diao,
        and~Dongyan~Wang*
\thanks{$^{\dag}$ These authors contributed equally to this work. }
\thanks{* means corresponding author. }
\thanks{Xin Chen, Hua Zhou, Yu Zhu and Dongyan Wang are with Midea Emerging Technology Center, San Jose, 95134, USA. Xin's email: chen1.xin@midea.com, Yu's email: zhu.yu@midea.com, Hua's email: hua.zhou@midea.com, and Dongyan's email: dongyan.wang@midea.com.}
\thanks{Liang Diao is with Midea Artificial Intelligence Research Institute, Shenzhen, Guangdong, 528311, P. R. China. email: liang.diao@midea.com.}
}

%
%

%

\maketitle

\begin{abstract}
In this paper, we introduce a new and challenging large-scale food image dataset called ``ChineseFoodNet", which aims to automatically recognizing pictured Chinese dishes.  Most of the existing food image datasets collected food images either from recipe pictures or selfie. In our dataset, images of each food category of our dataset consists of not only web recipe and menu pictures but photos taken from real dishes, recipe and menu as well. ChineseFoodNet contains over 180,000 food photos of 208 categories, with each category covering a large variations in presentations of same Chinese food. We present our efforts to build this large-scale image dataset, including food category selection, data collection, and data clean and label, in particular how to use machine learning methods to reduce manual labeling work that is an expensive process. We share a detailed benchmark of  several state-of-the-art deep convolutional neural networks (CNNs) on ChineseFoodNet. We further propose a novel two-step data fusion approach referred as ``TastyNet", which combines prediction results from different CNNs with voting method. Our proposed approach achieves top-1 accuracies of 81.43\% on the validation set and 81.55\% on the test set, respectively. The latest dataset is public available for research and can be achieved at \url{https://sites.google.com/view/chinesefoodnet/}.
\end{abstract}


\IEEEpeerreviewmaketitle

\section{Introduction}
\label{intro}

\IEEEPARstart{F}{ood} plays an essential role in everyone's lives, and the behaviour of diet and eating impacts everyone's health~\cite{mesas2012selected}. Underestimating food intake directly relates to diverse psychological implications~\cite{livingstone2003markers}. In recent years, photographing foods and sharing them on social networks have become a part of daily life. Consequently, several applications have been developed to record daily meal activities in personal food log system~\cite{aizawa2013invited}\cite{kawano2013real}\cite{beijbom2015menu}, which are employed to computer-aided dietary assessment~\cite{liu2016deepfood}, further usage preference experiments~\cite{aizawa2014comparative}\cite{takahashi2017estimation}, calorie measurement~\cite{pouladzadeh2014measuring}  and nutrition balance estimation~\cite{aizawa2013food}\cite{mezgec2017nutrinet}.  As one of user-friendly ways to input of the food log, automatic recognition of dish pictures gives rise of a research field of interest.

\begin{figure}[!htp]
	\centering
	\includegraphics[width=0.5 \textwidth, height=6.5cm]{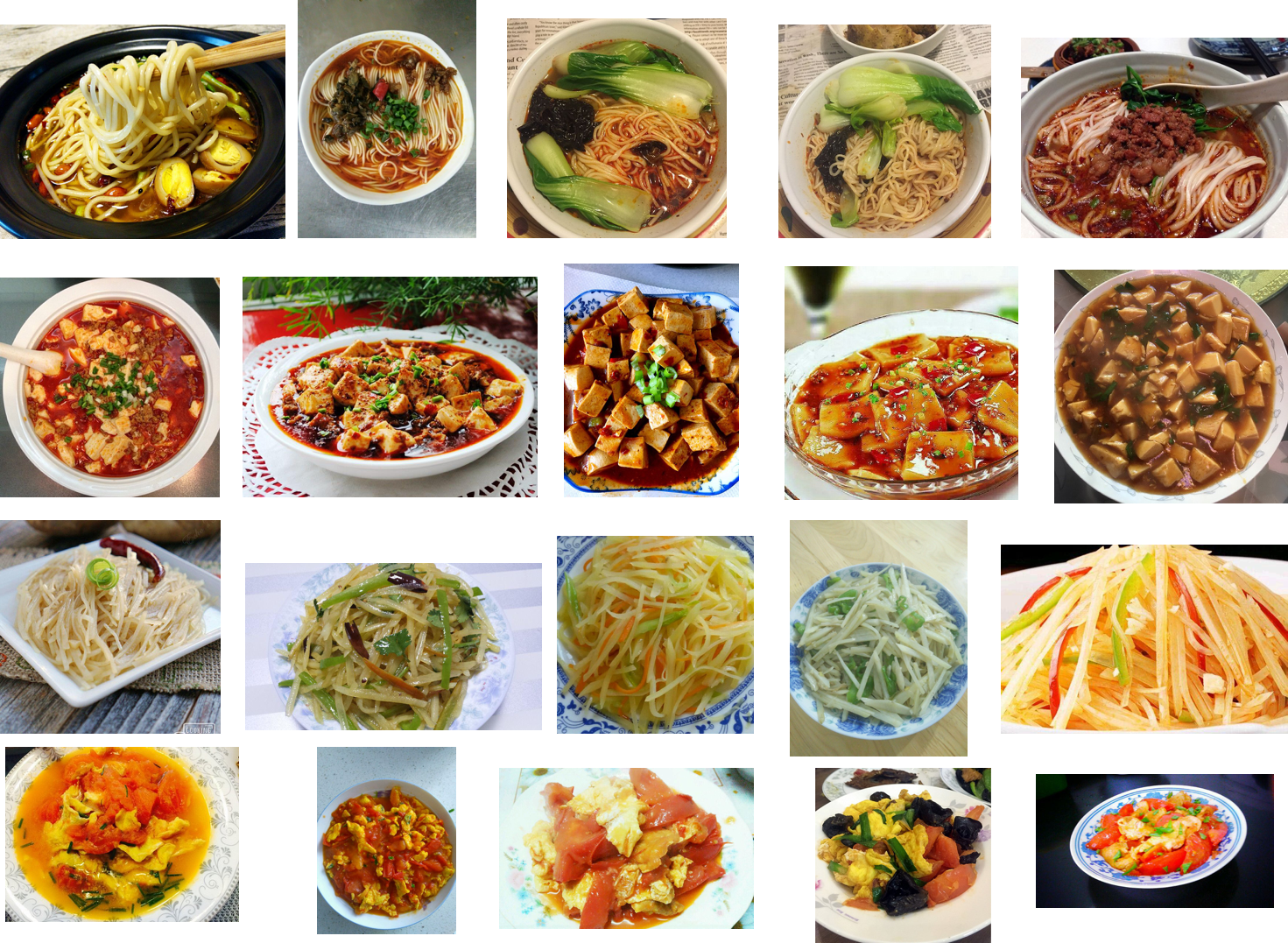}
	\caption{Example images from our dataset. Each row shows five images from one category of Chinese food. From top to bottom, the food names are Sichuan noodles with peppery sauce, Mapo tofu, potato silk, and scrambled egg with tomato, respectively. Variations in visual appearance of images of Chinese food caused by complex background, various illumination, different angle of view, different ingredients of the same category, etc. show challenges of visual food recognition. All of these image keep their original size.}
	\label{fig_food}
\end{figure}

Deep convolutional neural networks~(CNNs) have achieved state-of-the-art in a variety of computer vision tasks~\cite{lecun2015deep}\cite{schmidhuber2015deep}. The visual dish recognition task is the same situation~\cite{hassannejad2016food}. The quality of training datasets always plays an important role for training a deep neural network, where the high performance of the deep model is still data-driven to some extent \cite{bengio2009learning}\cite{deng2009imagenet}. 

However, to the best of our knowledge, there still exist no effective Chinese food recognition system matured enough to be used in real-world. The major reason is absence of large-scale and high quality image datasets. In~\cite{chen2012automatic}, the Chinese food dataset includes 50 categories, each of which has only 100 images. Obviously, the size of this dataset is not sufficient to satisfy deep learning training requirements. 

The visual dish recognition problem has widely been considered as one of challenging computer vision and pattern recognition tasks~\cite{hassannejad2016food}\cite{farinella2014benchmark}. Compared to other types of food such as Italian food and Japanese food, it is more difficult to recognize the images of Chinese dish as the following reasons:
\begin{enumerate}
\item The images of same category appear differently. Since most of the same Chinese dish have different ingredients and different cooking methods, the images are greatly visual different, even for human vision;
\item The noise of images of Chinese dishes is hard to model because of complex noise and a variety of backgrounds.The images of Chinese food are taken in various environment and complex background, for example dim light, vapour environment, strong reflection, various utensils of Chinese dishes such as color, shape, ornament,  etc. 
\end{enumerate}

\begin{figure*}[!htp]
	\centering
	\includegraphics[width=\textwidth, height=8.5cm]{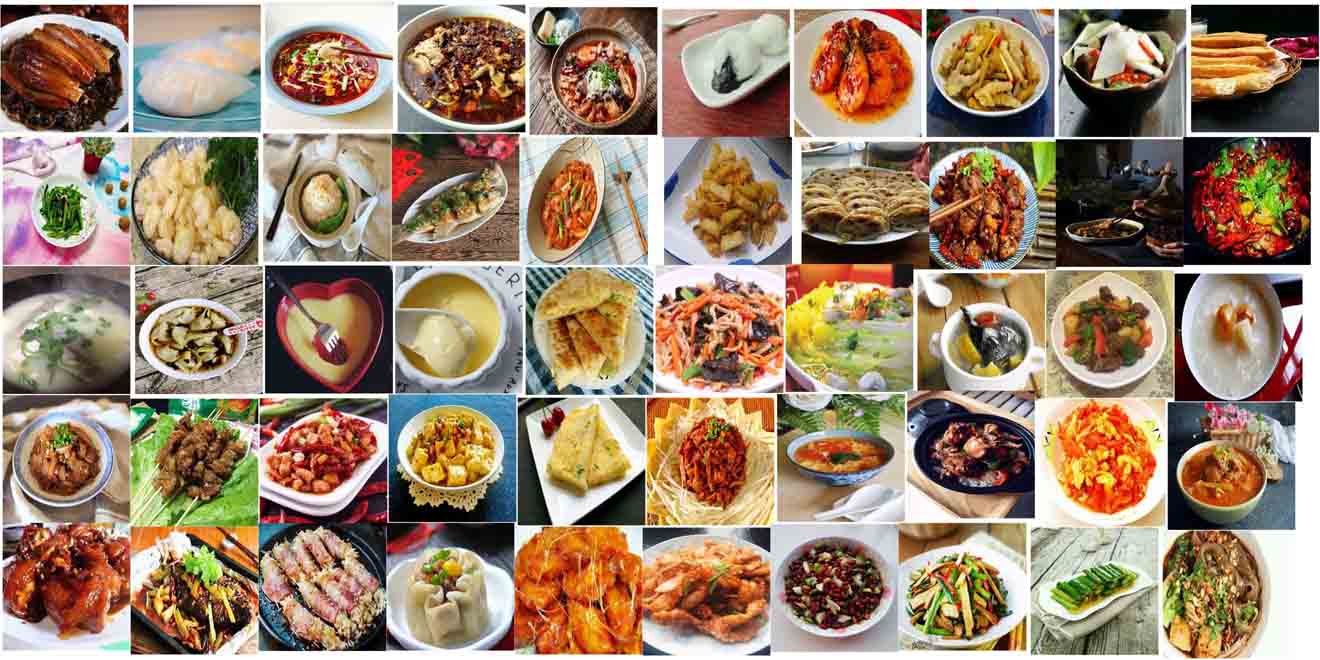}
	\caption{Fifty sample images of ChineseFoodNet dataset. The dataset contains 185,628 Chinese food images organized into 208 categories. All images in the dataset are color. Images are resized for better presentation}
	\label{fig_sim}
\end{figure*}

In order to give impetus to the progress of visual food classification and related computer vision tasks, we build a large-scale image dataset of Chinese dish, named by \textbf{ChineseFoodNet}. This dataset contains 185,628 images of 208 food categories covering most of popular Chinese food, and these images include web images and photos taken in real world under unconstrained conditions. To the best of our knowledge, ChineseFoodNet is the largest and most comprehensive dataset for visual Chinese food recognition. Some of images of ChineseFoodNet are shown in Figure.~\ref{fig_sim}.

We benchmark nine CNNs models of four state-of-the-art deep CNNs, SqueezeNet~\cite{iandola2016squeezenet}, VGG~\cite{simonyan2014very}, ResNet~\cite{he2016deep}, and DenseNet~\cite{iandola2014densenet}, on our dataset. Experimental results reveal that ChineseFoodNet is capable of learning complex models. 

In this paper, we also propose a novel two-step data fusion approach with voting. Although simple, voting is an effective way to fuse results~\cite{chen2016parallel}\cite{macdonald2006voting}. Guided by our benchmarks, we try some combination of different CNNs models Based on results on ChineseFoodNet, we take ResNet152, DenseNet121, DeneseNet169, DenseNet201 and VGG19-batch normalization~(BN)~\cite{ioffe2015batch} as our predictive models. \footnote{The name of CNNs networks consists of letters$+$numbers. Letters are type of CNNs, and following numbers are the number of layers.} Then we fusing these results with voting as a final result. This method is designated as`` \textbf{TastyNet}". Our proposed method has achieved top-1 accuracy 81.43\% in validation set and  81.55\% in test set, respectively. Compared to best results of the approaches with a single network structure, the improvements of 2.38\% in validation set and 2.33\% in these sets have been achieved, respectively.  

This paper takes three major contributions as following:
\begin{enumerate}
	\item We present a large-scale image dataset, ChineseFoodNet, for Chinese food recognition tasks. ChineseFoodNet is made up with 185,628 images of 208 categories, and most of the food image are from users' daily life. It is public available for research in related topics. \footnote{Our dataset can be accessed from \url{https://sites.google.com/view/chinesefoodnet/}.}
	\item We provide a benchmark on our dataset. Totally nine different models of four state-of-the-art CNNs architectures are evaluated. We presents the details of the methodology used in the evaluation and the pre-trained models will be public available for further research.
	\item We propose a novel two-step data fusion approach for visual food recognition, which combines predictive results of different CNNs with voting. Experimental results on ChineseFoodNet have shown that approach improves the performance compared to one deep CNNs model. It has shown that data fusion should be an alternative way to improve accuracy instead of only increasing numbers of layers in CNNs. 
	
\end{enumerate}

The paper is organized as follows. Section~\ref{related} briefly reviews some public food datasets and the state-of-the-art visual food recognition methods. Section~\ref{ChineseFoodNet} describes the procedure of building and tagging the ChineseFoodNet dataset. In section~\ref{benchmark}, several state-of-the-art CNNs methods are benchmarked on ChineseFoodNet. Section~\ref{tastynet} details our proposed data fusion approach and present our results on ChineseFoofNet. This paper closes with a conclusion of our work and some future directions in section~\ref{conclusion}.

\section{Related Work}
\label{related}
\subsection{Food Dataset}
The scholars have developed some public food datasets\footnote{In order to review fairly, we only discuss the data that are available for download in this paper. The last access date is June 1, 2017} for food-related applications such as dietary assessment, computational cooking, food recipe retrieval and so on. Pittsburgh Food Image Dataset (PFID) collects 4,556 fast food images~\cite{chen2009pfid}. The UNICT-FD889 dataset of 3,583 images related to 889 distinct dishes are used for Near Duplicate Image retrieval (NDIR)~\cite{farinella2014benchmark}. UEC-Food100~\cite{matsuda12_UECFOOD-100} and UEC-Food256~\cite{kawano14c_UECFOOD-256} are both Japanese food datasets and contain 100 and 256 categories, respectively. The UPMC-FOOD-101~\cite{wang2015recipe} and ETHZ-FOOD-101~\cite{bossard2014food} datasets are twin datasets and have same 101 food categories but different images. The images of UPMC-FOOD-101 are recipe images, in which each has the additional textual information, and the images of ETHZ-FOOD-101 are selfies. VIREO-172~\cite{chen2016deep} is a Chinese Food dataset containing a total of 353 ingredient labels and 110,241 images. However, it aims at cooking recipe retrieval with ingredient recognition. 
\subsection{Visual Dish Recognition}
Before introducing deep learning techniques to classification, traditional approaches with hand-crafted features have been applied to visual food recognition, including the pairwise feature distribution (PED)~\cite{yang2010food}, Gabor filters~\cite{zhu2008technology}, SIHT-based Bag of Visual Words (BoW)~\cite{kong2012dietcam}~\cite{kawano2013real}, optimized bag-of-features model~\cite{ anthimopoulos2014food}, co-occurrence~\cite{matsuda2012multiple}, textons~\cite{farinella2014classifying}, Random Forests (RF)~\cite{bossard2014food}, and Fisher Vector~\cite{kawano2014foodcam}. Like deep learning applied to other computer vision tasks, CNNs models have outperformed all of traditional methods and achieve higher and higher accuracy with deeper and deeper CNNs ~\cite{kawano2013real}\cite{liu2016deepfood}\cite{hassannejad2016food}\cite{christodoulidis2015food}\cite{martinel2016wide}\cite{kawano2014food}. 

However, all of these approaches of both traditional methods and deep learning haven't been tested on a large-scale image dataset of Chinese food. 

\section{ChineseFoodNet: a Large-scale Chinese Food Image Dataset}
\label{ChineseFoodNet}
To the best of our knowledge, there is no such large-scale image datasets for Chinese dish recognition which is mature enough to provided necessary resources for the data-driven techniques, e.g. deep learning, to train complex food recognition models. In this section, we present our procedures to build ChineseFoodNet. Labelling image is an expensive step in building large-scale dataset. In this paper, we design and develop a semi-supervised method to accelerate the whole process. 

\subsection{Category Selection}
\label{category_selction}
Various cooking styles exist in Chinese food culture, such as Sichuan cuisine, Canton cuisine, etc. Our Chinese food dataset must cover the most popular of Chinese cuisines from different styles of cooking.  In this subsection, we present our efforts to meet this goal.

First, 250 food categories are gathered from the internet.\footnote{\url{www.top.baidu.com}} However, some dishes are missed in search engine yet because they are too popular to be searched such as Tomato omelette. In order to cover them, we conduct a survey of favorite Chinese dishes within our group. Combining with results of the survey, we select about 300 categories. Since Chinese cruise categories is complex and some dishes are very similar visually, such as Braised Chicken Wings and Cola Chicken Wings, we manually merge related categories. After this process, 208 categories of Chinese dish are taken.\footnote{The names of Chinese dish in ChineseFoodNet are also listed at \url{https://sites.google.com/view/chinesefoodnet/}}

\subsection{Data Collection}
\label{data_collection}
There are two resources of our images, web images and taken photos. The web images in our dataset are coming from social network of the Chinese food and drink/cooking,\footnote{ \url{www.douguo.com}} where users uploaded their Chinese food pictures and also provided the tags (labels) of the image. Also some partial of the images in this dataset are collected by our group in daily life. 

After these steps, the number of images we brought together achieves more than 500,000. However, those images may contain missing labels, incorrect labels or unclear labels.
 
\subsection{Data Clean and Label}
\label{data_clean}
After collecting large number of food images, the next step is to clean these data and generate proper labels for each image. In this step, we first remove the images with irregular height or width (too large or too small) which usually are irrelevant images. Then we use entropy to clean the images without content. Entropy is a quantitative metric of image content~\cite{sonka2014image}. We calculate the value of entropy of each channel. If the value of any channel is small, we remove it because the image doesn't have enough useful information. The following step is to remove duplicate and/or very similar images with two steps. First, we calculate 1,024 deep features with the last full connection layer of AlexNet~\cite{krizhevsky2012imagenet}. Second, we calculate the Euclidean distance to measure the similarity. If the distance is below a threshold, we consider the images are very similar and remove one. 

Some of these images are clearly categorized with specific Chinese food name, such as most of recipe and menu images. The ground truth of this type of image can be directly extracted. However, the number of such images is very limited and the quality of those images are usually very high, e.g., the images are shot with sufficient light condition, good presentation of the food, and good angels, etc. Thus this type of images shows very different distributions comparing to the images captured in daily life, and brings a potential impact for the food recognition tasks in real life. 

The other images are usually not well-labeled, and the food photos are taken in real world conditions. Those images are mainly from the users' daily uploads which show very preferred data distributions in food images in the wild. Besides, this type of images is usually associated with metadata. The metadata can be viewed as an description of each image in text format, which often describes the name, cooking recipe and other information about the food in that image. In our procedure, this metadata is utilized to filter the useful images with correct labels. Particularly, we manually generate a set of keywords for each food class in our database, and use each set of the keyword to match the image metadata. Images with metadata which contains the keywords of certain class are selected and labeled with that class. 

It should be noted that, after the aforementioned step, there are still a number of incorrect labels, which are either caused by unclear descriptions in metadata or irrelevant images. Label validation by human labor on this large number of images is an expensive task in terms of both time and costs. Here we accelerate to label these image by some already labeled samples in advance. We first collect a small database of food images using the crowd-sourcing platform (no overlap between our current dataset) with same class labels. Then a shallow CNN model is trained for the food recognition task on this small database. Given this CNNs model, we classify our collected images into different classes representing candidate labels. Specifically, top $n$ (e.g., $5$) predictions from the shallow network are selected as the candidate labels for one image. Finally, we perform manually label validation to finalize the dataset by eliminating the wrong labelled images. 

\begin{figure*}[!htp]
	\centering
	\includegraphics[width=0.85\textwidth]{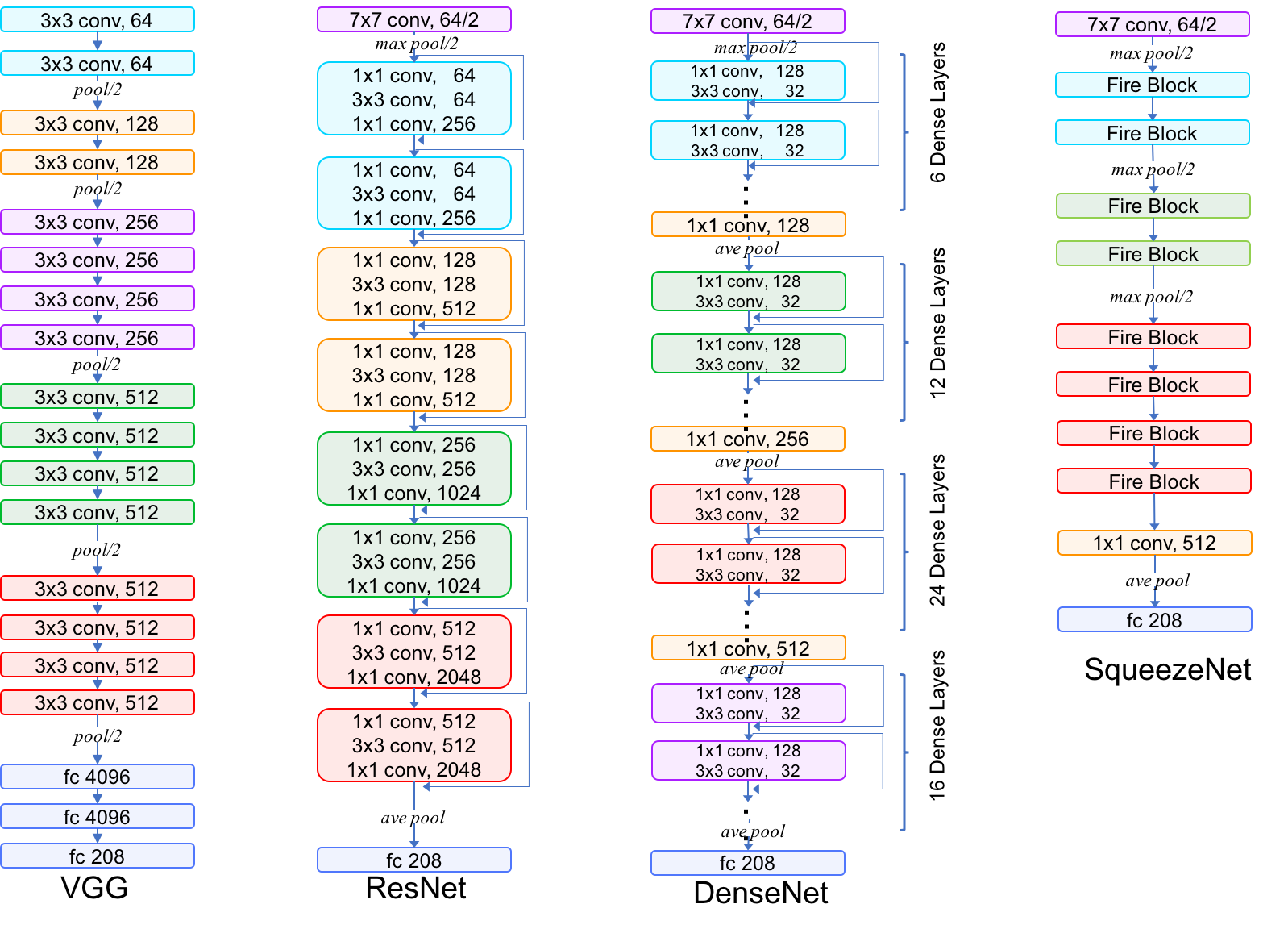}
	\caption{We show basic architectures of four well-known CNNs in our evaluation. From left to right, the architectures are VGG, Resnet, Densenet, and Squeezenet, respectively.}
	\label{fig:net}
\end{figure*}

\subsection{Dataset Description}
After work of category selection, data collection, the data collection and cleaning mentioned in previous subsections, finally the ChineseFoodNet dataset contains 185,628 images, with total size of 19.4 Gigabyte (GB). Images in the dataset are kept their original size without any processing and color. The total number of categories is 208 for the current version of dataset, and each image is labelled with only one label from 0 to 207. 

We split the whole dataset into training, testing and validation sets, approximately in the ratio 80\%, 10\% and 10\%, respectively. Specifically, there are 145,066, 20,254 and 20,310 images for training, validation and testing set, respectively.  Figure \ref{fig_food} and Figure \ref{fig_sim} show some example images in our dataset.

\section{Benchmark on ChineseFoodNet Dataset}
\label{benchmark}

In this section, we conducted benchmark experiments for the ChineseFoodNet dataset. First the experimental settings are described, then
we introduce the experimental protocol and finally we provide the
experimental results and analysis.

\subsection{Experimental settings}

Our experiments were all conducted using PyTorch~\cite{pytorch} deep learning framework.
In the training phase, the initial learning rate is set to 0.01, momentum
is set 0.9, and weight\_decay is set to 1e-4. We set the learning
rate to the initial learning rate decayed by 10 every 30 epoch. The
number of epoch for the training is set to 90. Training optimization
method is selected to stochastic gradient descent (SGD) with momentum. No augmentation process is applied except the resizing and mirror. Training images are firstly
resized to 256x256, then a random crop of size 224x224 with hoizontal
flip (probability 0.5) is applied. We have used the pretrained models
from imagenet dataset \cite{deng2009imagenet} and fine-tuned the network with our food data.
During the testing, images are resized to 256x256 and then we use
center crop of size 224x224 to feed into the network. All the experiments
were conducted on CentOS 7 operation system, with Intel Xeon E5 CPU (2.2G),
128GB RAM and Nvidia P100 Tesla GPUs hardware with 16G memory.

\subsection{Experimental Protocol}

\begin{table*}[!htp]
\centering
\caption{Recognition rates of different deep networks on our food
dataset. Both top-1 and top-5 accuracy are shown on validation set
and test set.}
\begin{tabular}{|l|c|c||c|c|}
\hline 
\multirow{2}{*}{\textbf{Method}} & \multicolumn{2}{c||}{\textbf{Validation}} & \multicolumn{2}{c|}{\textbf{Test}}\tabularnewline
\cline{2-5} 
 & \textbf{Top-1 Accuracy} & \textbf{Top-5 Accuracy} & \textbf{Top-1 Accuracy} & \textbf{Top-5 Accuracy}\tabularnewline
\hline 
\hline 
Squeezenet1\_1 & 58.42\% & 85.02\% & 58.24\% & 85.43\%\tabularnewline
\hline 
VGG19-BN & 78.96\% & 95.73\% & \textbf{79.22\%} & \textbf{95.99\%}\tabularnewline
\hline 
ResNet18 & 73.64\% & 93.53\% & 73.67\% & 93.62\%\tabularnewline
\hline 
ResNet34 & 75.51\% & 94.29\% & 75.82\% & 94.56\%\tabularnewline
\hline 
ResNet50 & 77.31\% & 95.20\% & 77.84\% & 95.44\%\tabularnewline
\hline 
ResNet152 & 78.34\% & 95.51\% & 79.00\% & 95.79\%\tabularnewline
\hline 
DenseNet121 & 78.07\% & 95.42\% & 78.25\% & 95.53\%\tabularnewline
\hline 
DenseNet169 & 78.87\% & \textbf{95.80\%} & 78.72\% & 95.83\%\tabularnewline
\hline 
DenseNet201 & \textbf{79.05\%} & 95.79\% & 78.78\% & 95.72\%\tabularnewline
\hline 
\end{tabular}
\label{table:1}
\end{table*}

\begin{figure}[!htp]
	\centering
	\includegraphics[width=0.35\textwidth]{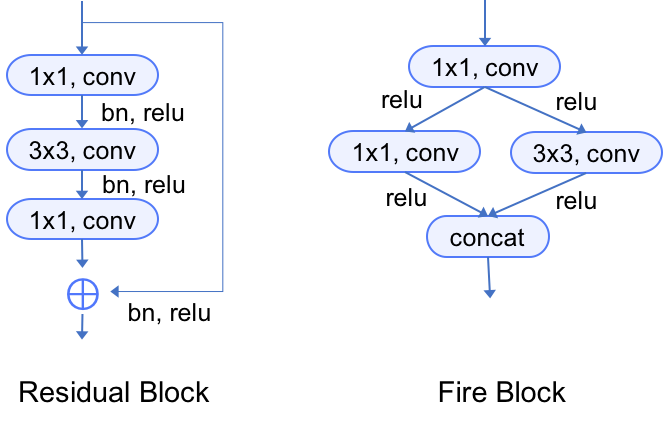}
	\caption{Illustration of residual block in ResNet and fire block in Squeezenet.}
	\label{fig:block}
\end{figure}

The dataset is split into training, validation, and test sets by
random selection. There are 145,065 images in the training set. There
are 20254 images in the validation set and the rest 20310 images are
used for testing. Comprehensive experiments have been conducted using
various popular deep learning network architectures with different
structures and different number of layers. Specifically, we have benchmarked
the performance of: Squeezenet (version 1.1) \cite{iandola2016squeezenet}, VGG19 (with BN
layer) \cite{simonyan2014very}, Resnet (18, 34, and 50) \cite{he2016deep}, DenseNet (121, 169,
and 201) \cite{iandola2014densenet}. In order to have a fair comparison, all the experiments
are using same input image size and same preprocessing/postprocessing procedures. Some implementation details of ResNet and Squeezenet are illuminated in Figure.~\ref{fig:block}. 

\subsection{Experimental Results}

The recognition performance of different deep networks are shown in
Table \ref{table:1}, both top 1 accuracy and top 5 accuracy are presented. Table \ref{table:1} has shown that the best top-1 performance
on validation set is 79.05\%, which is achieved by DenseNet201. The accuracies
of VGG19 and DenseNet169 are also very close to the best results.
On the test set, the best recognition rate is 79.22\%, obtained by
VGG19, the second best results is obtained by Resnet152, which is
0.22\% lower than then VGG19. 

Deeper CNNs models generally achieve better performance~\cite{szegedy2015going}\cite{szegedy2016rethinking}. From the results, we can see that, CNN models obtains significant improvements in performance when number of layers in same network architecture are increased. E.g., ResNet with 18 layers has recognition rate 73.64\%, while the deeper mode ResNet with 152 layers achieves about 5\% improvement in both validation and test sets. Similar results can be observed in DenseNet architecture. On the other hand, deep models with wider structure also shows promising performance, e.g., VGG19-BN obtains the best results in test set, and the worst result (58.42 \% and 58.24\% on validation and test sets, respectively) is achieved by Squeezenet v1.1, which is a shallow and narrow network structure designed for fast and efficient inference.  
\begin{figure*}[!htp]
	\centering
	\includegraphics[width=0.8\textwidth]{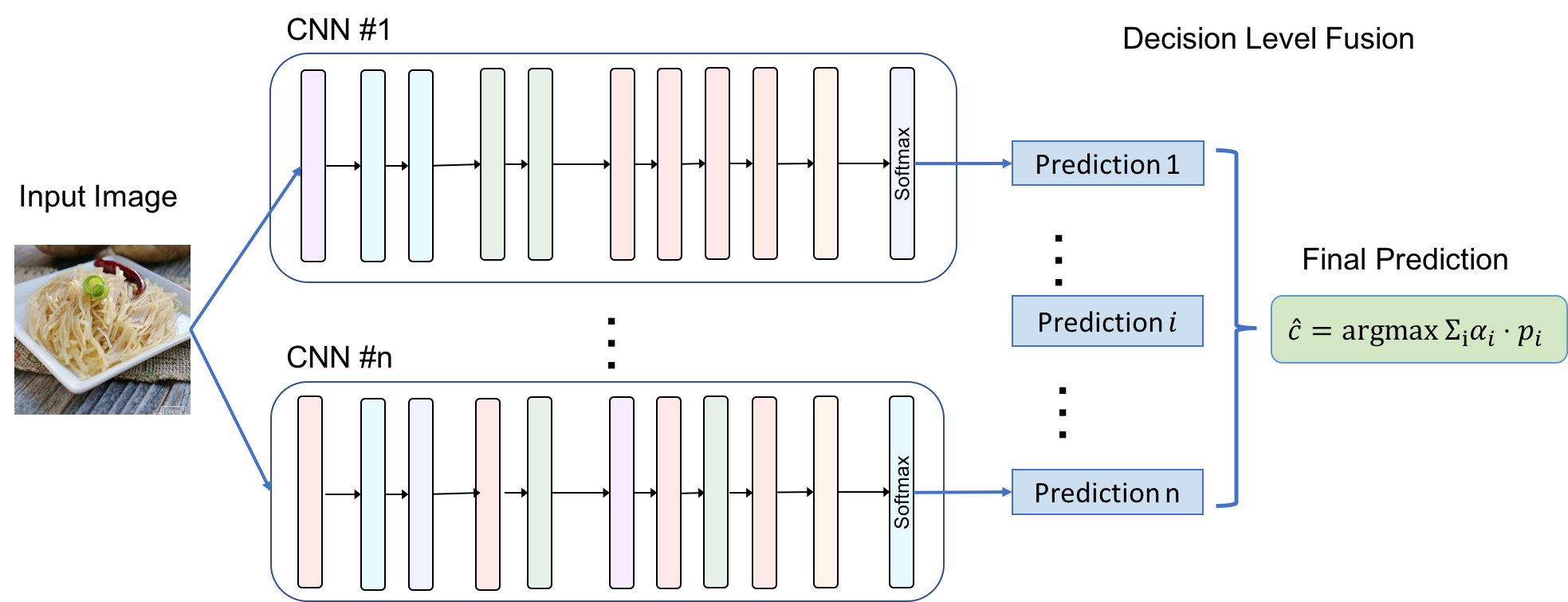}
	\caption{Basic scheme of the two-step data fusion approach. The first one is to obtain some predictive results from different models. In TastyNet, we use Resnet152,  DenseNet121, DenseNet169, DenseNet201 and VGG19-BN. The second one is to combine these result to one final result with voting policy. It his paper, we use weighted coefficient of the results of the first step.}
	\label{fig:flow}
\end{figure*}

\section{TastyNet: a two-step data fusion approach}
\label{tastynet}
\subsection{Methodology}
As shown in Table~\ref{table:1}, the accuracy has higher and higher with deeper and deeper model. If we would improve furthermore, a possible way to use much deeper CNNs models. However, it needs much computation and memory resources. What is more, deeper models easily lead to overfitting problem. The alternative way is the data confusion approach. Its idea is to fuse the inference results of different models. As shown in Figure.~\ref{fig:flow}, predictions
from different networks are gathered and a voting approach is utilized
to obtain the final fused prediction.

Based on some results of different combinations, as shown in Table.~\ref{table:2}, we select the combination of models that achieves the best top 1 result, ResNet152, DenseNet121, DenseNet169, DenseNet201 and VGG19-BN. The voting method is to average the results of all models. The algorithm is details in Algorithm.!\ref{algorithm:tastynet}.

\begin{algorithm}[!htbp]
	\caption{Algorithm of TastyNet.}\label{algorithm:tastynet}
	\begin{algorithmic}[1]
		\State Input:
		\State  Image 
		\State Output:
		\State Number \Comment{Range from 0-207}
		\State Predictive result from Resnet152, $p(i)$, $i$ from 0 to 207
		\State Predictive result from DenseNet121, $p(i)$, $i$ from 0 to 207
		\State Predictive result from DenseNet169, $p(i)$, $i$ from 0 to 207
		\State Predictive result from DenseNet201, $p(i)$, $i$ from 0 to 207
		\State Predictive result from Resnet152, $p(i)$, $i$ from 0 to 207
		\State Get average result $\overline{p}(i)$ of all $p(i)$, $i$ from 0 to 207
		\State Find maximum $\overline{p}(i)$ and get $i$
		\State The output is number $i$
	\end{algorithmic}
\end{algorithm}

\begin{table*}
\centering
\caption {Experimental results (recognition accuracies) of different fusion schemes}
\begin{tabular}{|l|c||c||c|c|}
\hline 
\multirow{2}{*}{\textbf{Fusion Method}} & \multicolumn{2}{c||}{\textbf{Top 1 Accuracy}} & \multicolumn{2}{c|}{\textbf{Top 5 Accuracy}}\tabularnewline
\cline{2-5} 
 & \textbf{Validation} & \textbf{Test} & \textbf{Validation} & \textbf{Test}\tabularnewline
\hline 
\hline 
ResNet (18 + 34 + 50 + 152) & 79.19\% & 79.46\% & 96.03\% & 96.16\%\tabularnewline
\hline 
DenseNet (121 + 169 + 201) & 80.47\% & 80.17\% & 96.26\% & 96.30\%\tabularnewline
\hline 
ResNet (18 + 34 + 50 + 152) + Densenet (121 + 169 + 201) & 80.89\% & 81.08\% & 96.60\% & 96.67\%\tabularnewline
\hline 
ResNet (18 + 34 + 50 + 152) + Densenet (121 + 169 + 201) + VGG19-BN & 81.23\% & 81.12\% & \textbf{96.79\%} & \textbf{96.76\%}\tabularnewline
\hline 
ResNet152 + DenseNet (121 + 169 + 201) + VGG19-BN & \textbf{81.43\%} & \textbf{81.55\%} & 96.73\% & \textbf{96.76\%}\tabularnewline
\hline 
\end{tabular}\label{table:2}
\end{table*}
\subsection{Results and Analysis}
 Different combinations of network architectures are applied and the experimental results are shown in Table \ref{table:2}. From this table, we can conclude that the overall performance is generally increasing for different combinations with ensemble more deep networks. The fusion results of ResNet with different number of layers, obtained higher performance (79.46\% top 1 accuracy on test set) than single ResNet (ResNet 152, 77.84\% top 1 accuracy on test set). Also the fusion results on DenseNet achieved a 1.12\% improvement on test set than the best results achieved for single DenseNet architecture. Furthermore, combination of different types of CNNs networks (e.g., ResNets, DenseNets and VGG shown in Row 3 and 4 in Table \ref{table:2}) further improves the overall recognition performance. The best result is obtained by fusing
Resnet152 and Densenet 121, Densenet 169, Densenet 201, and VGG19-BN,
the recognition accuracy is 81.43\% on the validation set and 81.55\%
for the test set. This results is 2.38\% and 2.33\% higher than the
single network on validation and test set, respectively. Based on the experimental results, we select five CNNs models ,Resnet152, DenseNet121, DenseNet169, DenseNet201 and VGG19-BN, as components of TastyNet. 

From our proposed approach, we get two conclusions as followings:

\begin{enumerate}
	\item By applying data fusing approach on different deep networks,
	the overall performance can be further boosted than using the single
	deep network;
	\item Combination of different network architectures show more benefits in improving the performance than the combinations with same network architectures, and combination of deeper and wider networks obtains the best results in our evaluation;
	
\end{enumerate}

\section{Conclusion and Future Work}
\label{conclusion}
In this paper, we have successfully created a very large-scale image dataset for Chinese dish recognition, ChineseFoodNet. It contains 185,628 images of 208 food categories, in which the images are from not only web images but also real world. As a consequence, the models trained on our dataset should have covered most of food recognition applications.  Also, we present the benchmarks of nine state-of-the-art CNNs models of four well-known CNNs architectures on ChineseFoodNet.  Finally, we propose a novel two-step data fusion approach, "TastyNet". Based on experimental results, we select Resnet 152, Densenent 121, Densenet 169, Densenet 201 and VGG19$+$BN models. After voting the results of these model, we obtain final inference result. It has shown the state-of-the-art results on ChineseFoodNet. What is more, our proposed approach has shown that data fusion is an effective way to obtain a better result instead of only working on one type CNNs model.

For our future work, we are extending the number of food category to over 500 that should be applied in much applications. Also, we will investigate new fusion methods to fuse the different results with different models to obtain the better performance.

\section*{Acknowledgment}

The authors would like to thank Zhi Zhang, Xukai Zhang, Zigang Wang, Xiangping Zeng, Xiaofei Xu, Dangdang Mi and Qingqing Chang for their discussions and efforts to collect data. All of them are with Midea Health and Nutrition Institute.

\ifCLASSOPTIONcaptionsoff
  \newpage
\fi

\bibliographystyle{IEEEtran}
\bibliography{foodnet}

%

\end{document}